# Economic Analysis and Optimization of Energy Storage Configuration for Park Power Systems Based on Random Forest and Genetic Algorithm

Yanghui Song#, Aoqi Li#, Lilei Huo#

*School of Information Science and Technology, Yunnan Normal University, Kunming, 650500, China*
*#These authors contributed equally.*

***Abstract:*** *This study aims to analyze the economic performance of various parks under different conditions, particularly focusing on the operational costs and power load balancing before and after the deployment of energy storage systems. Firstly, the economic performance of the parks without energy storage was analyzed using a random forest model. Taking Park A as an example, it was found that the cost had the greatest correlation with electricity purchase, followed by photovoltaic output, indicating that solar and wind power output are key factors affecting economic performance. Subsequently, the operation of the parks after the configuration of a 50kW/100kWh energy storage system was simulated, and the total cost and operation strategy of the energy storage system were calculated. The results showed that after the deployment of energy storage, the amount of wind and solar power curtailment in each park decreased, and the operational costs were reduced. Finally, a genetic algorithm was used to optimize the energy storage configuration of each park. The energy storage operation strategy was optimized through fitness functions, crossover operations, and mutation operations. After optimization, the economic indicators of Parks A, B, and C all improved. The research results indicate that by optimizing energy storage configuration, each park can reduce costs, enhance economic benefits, and achieve sustainable development of the power system.*

## 1. Introduction

The microgrid in the park is powered by both renewable energy sources like wind and solar, as well as the main grid, to jointly supply power to the loads. To maximize the proportion of renewable energy in the load and minimize the problem of power curtailment caused by the mismatch between the load and the power generation of wind and solar, this study aims to design and optimize the coordinated configuration of wind, solar, and storage in the park microgrid[1]. Corresponding configuration schemes are proposed to ultimately achieve maximized benefits and minimized costs. This paper investigates the economic performance and power load balancing of the park's power system under different energy storage configurations, especially analyzing and optimizing the energy storage system of each park through methods such as decision forests and genetic algorithms[2][3]. The goal is to reduce operational costs and enhance the economic efficiency and reliability of the system.

## 2. Energy storage analysis of each park

Consider three microgrid parks, each independently connected to the main grid. The installed capacities for photovoltaic (PV) and wind power, as well as the maximum load parameters for each park, are depicted in (Figure 1.). Here, $P_{pv}.A$ and $P_{pv}.C$ represent the PV installed capacities for Parks A and C, respectively, while $P_w.B$ and $P_w.C$ denote the wind power installed capacities for Parks B and C. Additionally, $P_{L_{max}}.A$, $P_{L_{max}}.B$, and $P_{L_{max}}.C$ are the maximum load values for Parks A, B, and C, respectively.

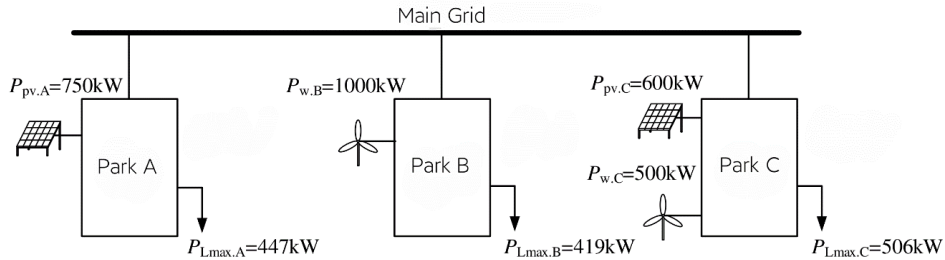

Figure 1: Microgrid Parks Connected to Main Grid Structure

## 2.1. Economic Analysis of Park Operations without Energy Storage

The paper employs a decision forest regression model to calculate and analyze the correlations between costs and various factors such as electricity purchases, photovoltaic (PV) or wind power generation, and the curtailment of wind and solar energy. For regression tasks, the prediction of a random forest is obtained by averaging the predictions of all the individual decision trees. In the text, let $T$ represent the number of decision trees in the forest, and $h_j(x)$ denote the prediction of the $jth$ tree for the sample $x$. Thus, the prediction $\hat{y}$ of the random forest and regression expressions are presented can be expressed as:

$$\hat{y} = \frac{1}{T}\sum_{j=1}^{T} h_j(x) \tag{1}$$

$$y = \beta_0 + \beta_1 X_1 + \beta_2 X_2 + \cdots + \beta_n X_n + \varepsilon \tag{2}$$

where $y$ is the dependent variable, that is, the predicted variable, in this case, the hourly cost (CNY) of park A; $X_1 + X_2 + \cdots + X_n$ are independent variables, that is, variables used to predict dependent variables, in this case, they are photovoltaic output (p.u.), park A's purchased electricity (kW), and park A's curtailment (kW); $\beta_0$ is the intercept, which is the value of the dependent variable when all independent variables are zero; $\beta_1$, $\beta_2, \ldots, \beta_n$ are regression coefficients that indicate the degree to which each independent variable affects the dependent variable. $\varepsilon$ is the error term and represents random variation that the model failed to explain.

## 2.2. Economic Analysis of Parks with a 50kW/100kWh Energy Storage System

This paper simulates and analyzes the economic performance and operation of energy systems in each park equipped with a 50kW/100kWh energy storage system, including wind power generation, solar power generation, and energy storage. The storage configuration employs lithium iron phosphate batteries with a power unit cost of 800 CNY/kW and an energy unit cost of 1800 CNY/kWh. The State of Charge (SOC) operational range is set from 10% to 90%, with a charging/discharging efficiency of 95% and an operational lifespan of 10 years. An objective function is constructed and solved to obtain the economic indicators for each park. By adjusting input parameters and optimizing operational strategies, the paper demonstrates that it is possible to further enhance the economic benefits and reliability of the system[4].

### 2.2.1. Calculation of Energy Storage System Costs

The power cost ($C_P$) of the energy storage system:

$$C_P = P_S \times 800 \tag{3}$$

where $P_S$ represents the power parameters of the energy storage system.

The energy cost of the energy storage system:

$$C_e = E_s \times 1800 \tag{4}$$

where $E_S$ represents the energy parameters of an energy storage system.

The total cost of the energy storage system:

$$C_{all} = \frac{C_P + C_e}{L \times 365 \times 24} \tag{5}$$

where $L$ represents the load volume.

*2.2.2. Energy Storage Operation Strategy*

$$n_l = L - S - W \tag{6}$$

where $n_l$ represents the net load, $L$ represents the load, $S$ represents the photovoltaic power generation, and $W$ represents the wind power generation.

*2.2.3. If the net load is greater than 0, then discharging is required*

$$dch = min(n_l, P_s, (SOC - SOC_{min})) \tag{7}$$

$$SOC = SOC - dch \times \eta \tag{8}$$

$$n_l = n_l - dch \tag{9}$$

where $dch$ stands for discharge and $\eta$ represents charge and discharge efficiency.

*2.2.4. If the net load is less than 0, then charging is required*

$$dch = min(-n_l, P_s, (SOC_{max})) \tag{10}$$

$$SOC = \frac{SOC + ch}{\eta} \tag{11}$$

$$n_l = n_l + ch \tag{12}$$

where $ch$ stands for the amount charged.

*2.2.5. Calculation of Total Power Supply Cost*

After calculating the grid electricity cost ($g_c$), wind energy cost ($w_c$), solar energy cost ($s_c$), and energy storage system operation cost ($storage_{cost}$), the total power supply cost ($T_{cost}$) would be:

$$T_{cost} = g_c + w_c + s_c + storage_{cost} \tag{13}$$

*2.2.6. The total power supply cost without energy storage configuration*

To separately determine the grid electricity cost ($gcno_s$), wind energy cost ($wcno_s$), and solar energy cost ($scno_s$) without energy storage, the total power supply cost would be:

$$tcno_s = gcno_s + wcno_s + scno_s \tag{14}$$

there are the following constraint conditions:

(1) Charge and discharge efficiency of the energy storage system ($\eta$):

$$\eta = 95\% \tag{15}$$

(2) SOC limitations of the energy storage system:

$$10\% Capacity \leq SOC \leq 90\% Capacity \tag{16}$$

*2.3. Optimization of Energy Storage Size for Each Park*

This paper conducts a simulation of the application of energy storage in microgrids, where a genetic algorithm is used to find the optimal energy storage configuration[5]. The simulation incorporates a fitness function, crossover operation, mutation operation, energy storage operation strategies, as well as mathematical models and formulas for total power supply cost[6].

*2.3.1. Fitness Function*

The fitness function is used to evaluate the fitness of each individual, which is the total power supply cost.

$$f(s_p, s_c) = (s_p \times 100) + (s_c \times 100) \tag{17}$$

This function is used in the genetic algorithm to select and compare different energy storage configurations.

*2.3.2. Single-point Crossover Operation*

$$child_1 = \left( \alpha \times pt_{1_p} + (1-\alpha) \times pt_{2_p}, \alpha \times pt_{1_c} + (1-\alpha) \times pt_{2_c} \right) \tag{18}$$

$$child_2 = \left((1-\alpha) \times pt_{1_p} + \alpha \times pt_{2_p}, (1-\alpha) \times pt_{1_c} + \alpha \times pt_{2_c}\right) \quad (19)$$

where $\alpha$ is a random number, $pt_1$ and $pt_2$ are the selected parent individuals.

*2.3.3. Mutation Operation*

$$n_{individual} = (new_p, individual_c) or (individual_p, new_c) \quad (20)$$

where $new_p$ and $new_c$ are randomly generated new values within the defined ranges for energy storage power and capacity.

*2.3.4. Energy Storage Operation Strategy*

The operation strategy of the energy storage system involves calculating the net load and the change in SOC (State-of-Charge). When the load exceeds the generated power, the energy storage system discharges; and when the generated power exceeds the load, the energy storage system charges.

The formulas are as follows:

$$dch = min\left(n_l, P_s(SOC - SOC_{min})\right) \quad (21)$$

$$ch = min\left(-n_l, P_s, \frac{(SOC + ch)}{\eta}\right) \quad (22)$$

**3. Results**

*3.1. Economic Performance of Parks without Energy Storage*

Taking Park A as an example, we can see that (Figure 2.) shows that cost has the greatest correlation with electricity purchases, and at the same time, it is inevitable that it has a low correlation with curtailment. In addition to the inevitable relationship of purchased electricity, the correlation between photovoltaic output is high, so this paper can conclude that the key factors affecting the economy are photovoltaic and wind power output[7].

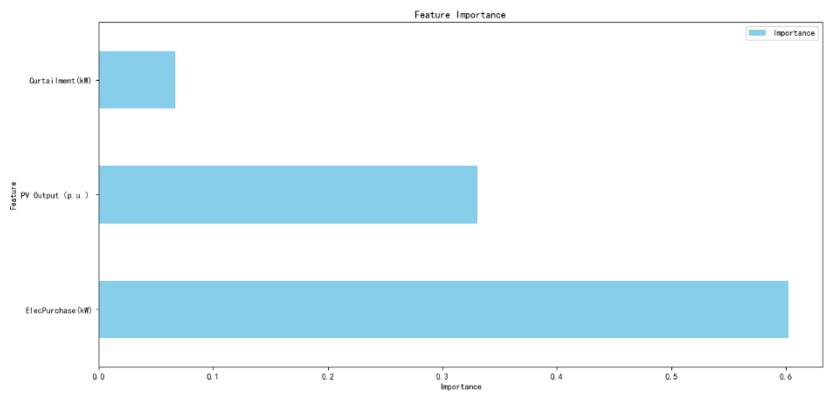

*Figure 2: Histogram of the correlation of economic indicators in park A*

The following diagrams are illustrations of the electrical load balance of Park A, Park B, and Park C without energy storage.

As can be seen from (Figure 3.), the load, heat supply, and electricity consumption of Park A show a certain regularity throughout the day. Starting in the morning, the load and heat supply gradually increase, reaching a peak and then starting to decline, and the electricity consumption also has a significant increase in the morning, and then a relatively low level in the afternoon and evening. Power generation is higher at night and lower during the day. From (Figure 4.), we can see the trend of the total power of Park B: the overall fluctuation is small, fluctuating around 350 kW; The change trend of the total traffic in the park: large fluctuations; The trend of power consumption in the park: the fluctuation range is less than the total power and total flow of the park, but there are also peaks and valleys. Trend of total load in the park: Similar to the trend of total power in the park, but slightly higher than the total power in the park. It can be seen from (Figure 5.) that the overall C load of the park decreases first and then increases.

The curtailment power of network wind is higher from 00:00:00 to 08:00:00, and then gradually decreases. The purchased electricity is low from 00:00:00 to 09:00:00, and then gradually increases; Utility electricity consumption is relatively stable and does not fluctuate much.

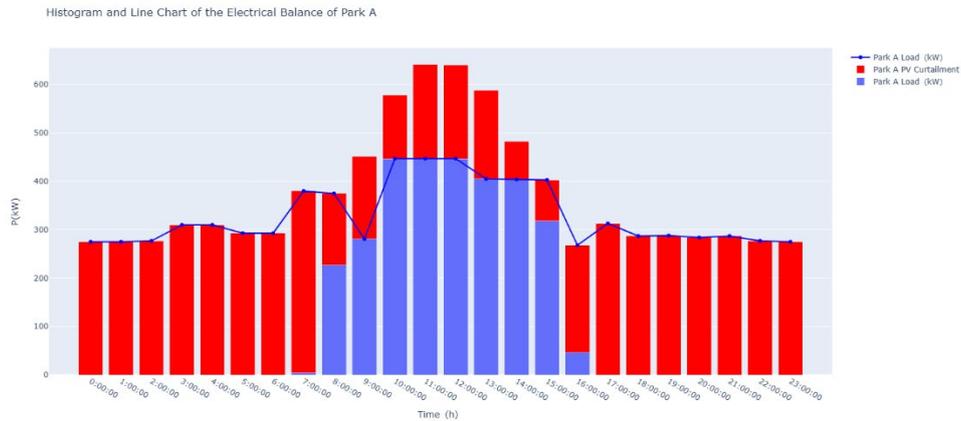

Figure 3: Histogram of the electrical load balance of Park A in the absence of energy storage

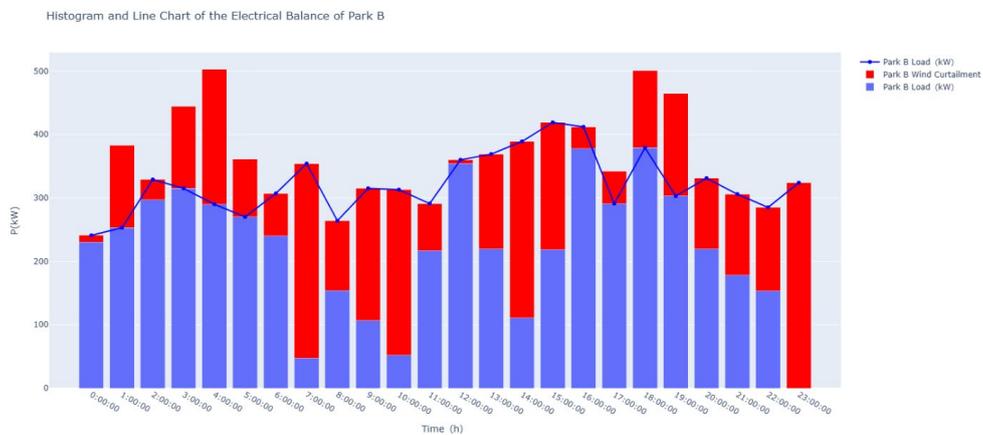

Figure 4: Histogram of the electrical load balance of Park B in the absence of energy storage

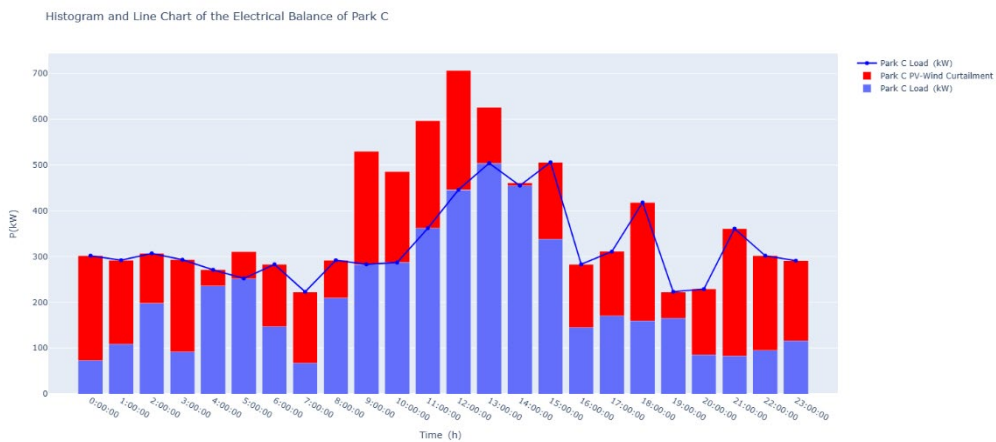

Figure 5: Histogram of the electrical load balance of Park B in the absence of energy storage

### 3.2. The Economic Analysis Results of Parks with a 50kW/100kWh Energy Storage System

Table. 1 compares the operation without energy storage configuration and with energy storage configuration. It can be seen that after the implementation of energy storage, the amount of wind and solar curtailment has been reduced, which can lower the operating costs of various parks.

*Table 1: Comparison of Parks with and without Energy Storage Configuration*

| Economic Indicators (50Kw/100kWh) | Park A | Configuration A | Park B | Configuration B | Park C | Configuration C |
|---|---|---|---|---|---|---|
| Electricity Purchased (kWh/day) | 4874.13 | 4747.81 | 2432.30 | 2220.78 | 2699.39 | 2517.67 |
| Wind and Solar Energy Waste (kW/day) | 951.20 | 875.20 | 897.50 | 744.60 | 1128.02 | 993.27 |
| Total Power Supply Cost (CNY/day) | 5609.27 | 5543.23 | 5071.15 | 4560.90 | 4960.79 | 4335.42 |
| Average Unit Cost of Electricity Supply (CNY/kW·h) | 0.81 | 0.70 | 0.65 | 0.58 | 0.65 | 0.56 |

Table. 2 presents the outcomes of energy storage optimization for parks A, B, and C. It is clear that while preserving the fluctuation patterns of wind, solar, and load power, the 50kW/100kWh configuration is suboptimal. Upon optimization employing a genetic algorithm, there has been an enhancement in the economic metrics across parks A, B, and C. Consequently, this manuscript posits that energy storage strategies ought to be custom-designed for individual parks through analytical assessment. The customized approaches, as illustrated by the enhancements achieved in this study, possess the potential to minimize expenditures and augment the earnings for each respective park.

*Table 2: Optimized Energy Storage Statistics for Each Park*

| Economic Indicators | OptimizationA (40Kw/140kWh) | Optimization B (50Kw/150kWh) | OptimizationC(60Kw/100kWh) |
|---|---|---|---|
| Electricity Purchased (kWh/day) | 4621.49 | 2136.57 | 2356.40 |
| Wind and Solar Energy Waste (kW/day) | 799.2 | 687.6 | 902.07 |
| Total Power Supply Cost (CNY/day) | 5461.85 | 4500.35 | 4235.51 |
| Average Unit Cost of Electricity Supply (CNY/kW·h) | 0.69 | 0.58 | 0.54 |

The following figures (Figure 6.-Figure 8.) show the column line graph of the change in SOC (State of Charge) over time for the three parks A, B, and C after the optimization of the energy storage system. It can be observed from the figure that the change trend of SOC is reasonable: SOC will change accordingly when it is necessary to abandon electricity or light, or when wind power and photovoltaic power start to generate electricity. Specifically, when the energy storage system is charging, the SOC curve shows an upward trend; while during the discharge process, the SOC curve decreases. This change trend effectively reflects the operation status of each park after the optimization of the genetic algorithm, indicating the effectiveness of the optimization strategy.

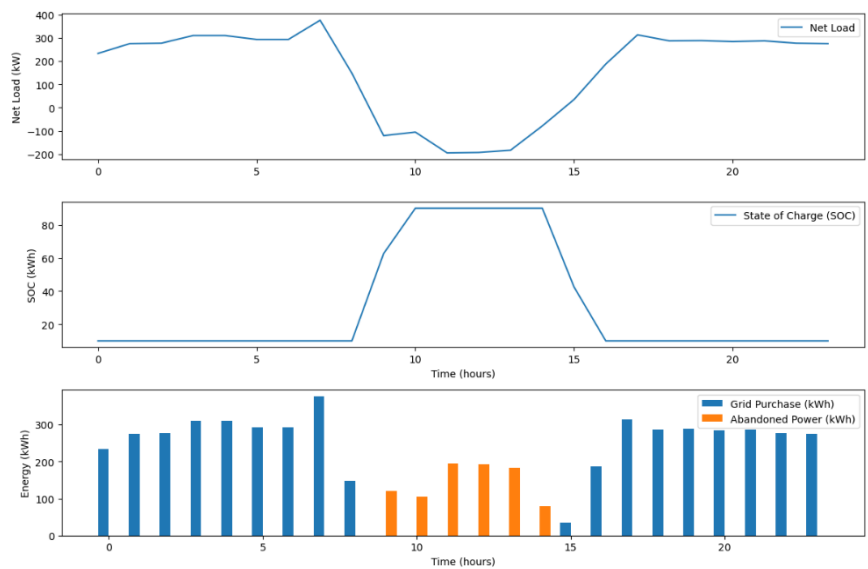

*Figure 6: Schematic of Charging and Discharging in Park A*

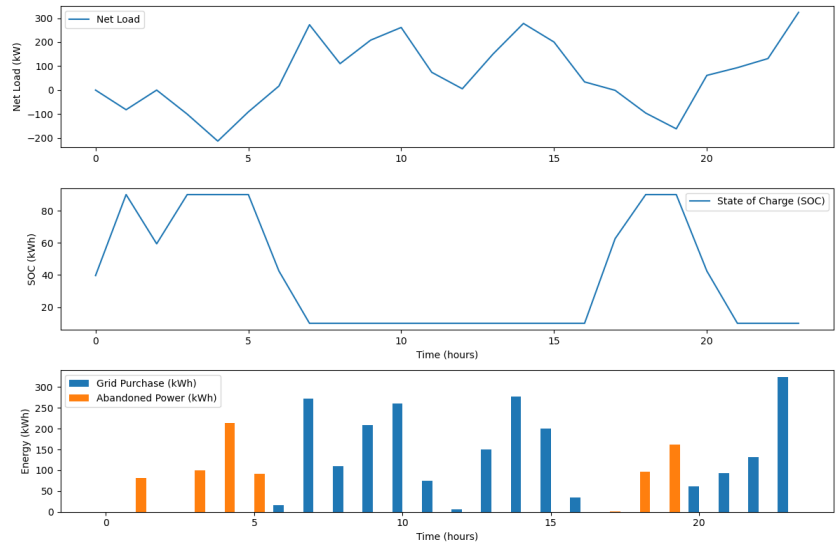

Figure 7: Schematic of Charging and Discharging in Park B

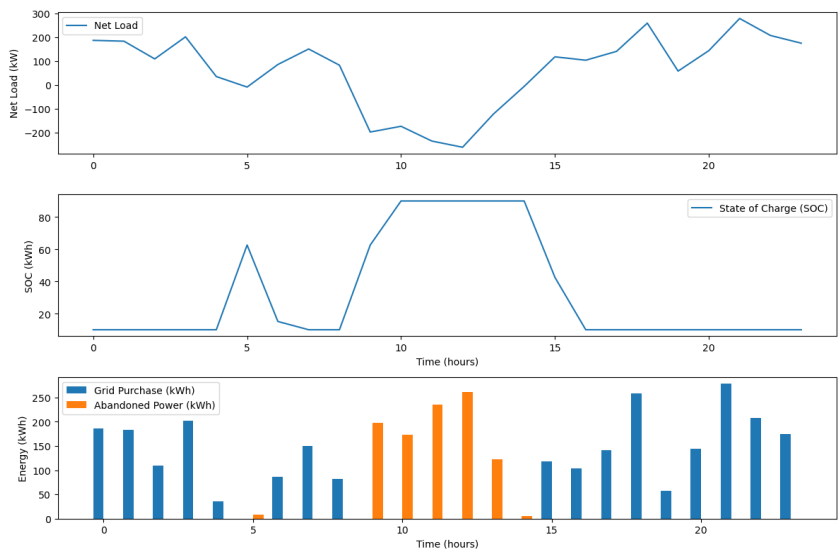

Figure 8: Schematic of Charging and Discharging in Park C

## 4. Conclusions

In this study, we conducted a timely and systematic review of the research efforts, utilizing a random forest model to analyze and optimize energy storage configurations within park power systems. Simulations integrating a 50kW/100kWh energy storage system demonstrated reduced curtailment and lowered operational costs, highlighting the benefits of energy storage. Further advanced optimization using a genetic algorithm brought economic benefits to individual parks. We concluded that data-driven, customized energy storage strategies are essential for economic optimization and sustainable development of power systems, providing a valuable framework for the integration of renewable energy in park power systems.